\documentclass[runningheads]{llncs}

\usepackage{eccv}

\usepackage{eccvabbrv}

\usepackage{graphicx}
\usepackage{booktabs}
\usepackage{multirow}

\usepackage[accsupp]{axessibility}  

\usepackage{hyperref}

\usepackage{orcidlink}

\begin{document}

\title{DVLO: Deep Visual-LiDAR Odometry with Local-to-Global Feature Fusion and Bi-Directional Structure Alignment} 

\titlerunning{DVLO for Multi-modal Odometry}

\author{Jiuming Liu\inst{1}\orcidlink{0009-0001-8047-3814} \and Dong Zhuo\inst{1}\orcidlink{0009-0005-6747-7481} \and Zhiheng Feng\inst{1}\orcidlink{0000-0001-8014-6410} \and
Siting Zhu\inst{1}\orcidlink{0009-0009-9521-2682} \\ Chensheng Peng\inst{2}\orcidlink{0000-0001-9213-5970} \and Zhe Liu\inst{3}\orcidlink{0000-0001-6753-0303} \and Hesheng Wang\inst{1}\orcidlink{0000-0002-9959-1634}\thanks{Corresponding author. The first two authors contribute equally.}}

\authorrunning{J. Liu et al.}

\institute{Department of Automation, Shanghai Jiao Tong University \and
University of California, Berkeley \and
MoE Key Lab of Artificial Intelligence, Shanghai Jiao Tong University \\
\email{\{liujiuming, paryi555, wanghesheng\}@sjtu.edu.cn} }

\maketitle

\begin{abstract}
Information inside visual and LiDAR data is well complementary derived from the fine-grained texture of images and massive geometric information in point clouds. However, it remains challenging to explore effective visual-LiDAR fusion, mainly due to the intrinsic data structure inconsistency between two modalities: Image pixels are regular and dense, but LiDAR points are unordered and sparse. To address the problem, we propose a local-to-global fusion network (DVLO) with bi-directional structure alignment. To obtain locally fused features, we project points onto the image plane as cluster centers and cluster image pixels around each center. Image pixels are pre-organized as pseudo points for image-to-point structure alignment. Then, we convert points to pseudo images by cylindrical projection (point-to-image structure alignment) and perform adaptive global feature fusion between point features and local fused features. Our method achieves state-of-the-art performance on KITTI odometry and FlyingThings3D scene flow datasets compared to both single-modal and multi-modal methods. Codes are released at \url{https://github.com/IRMVLab/DVLO}.
\end{abstract}

\section{Introduction}
\label{sec:intro}

Visual/LiDAR odometry is a fundamental task in the field of computer vision and robotics, which estimates the relative pose transformation between two consecutive images or point clouds. It is widely applied in autonomous driving \cite{jiang2024neurogauss4d, wang2021pwclo,zhu2024semgauss,wu2024emie}, SLAM \cite{deng2024compact, zhu2024sni, deng2024plgslam, wu2024dvn}, navigation \cite{10486967,liutnnls,10560465} etc. Recently, the multi-modal odometry \cite{graeter2018limo, wang2021dv, zhuoins20234drvo} has gained increasing attention because it can take advantage of complementary information from different modalities and possess the strong robustness for asymmetric sensor degradation \cite{deng2023long}. 

\begin{figure}[tb]
  \centering
  \includegraphics[width=1\linewidth]{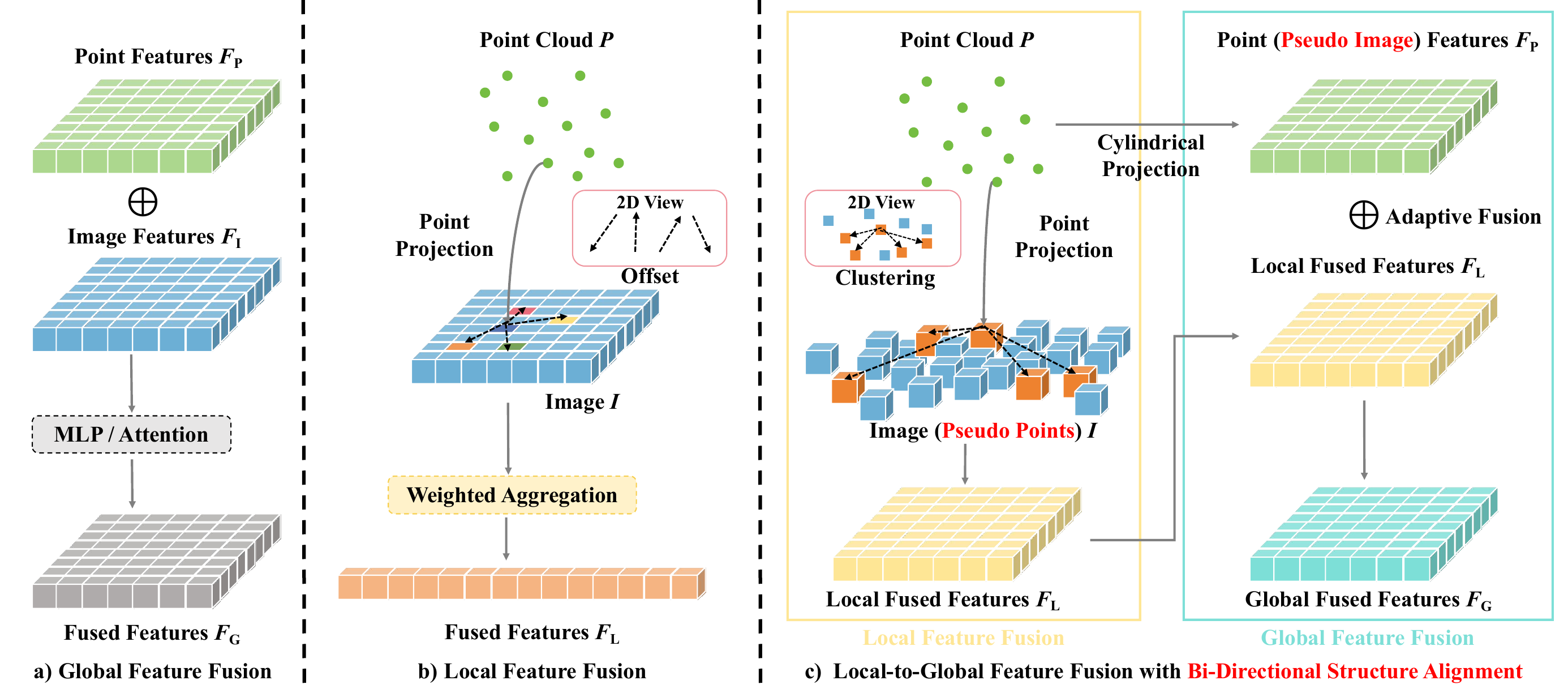}
  \caption{Different fusion strategies for images and points. Most previous works only perform the fusion globally \cite{valente2019deep} or locally \cite{zhuoins20234drvo}. Our DVLO designs a local-to-global fusion strategy that facilitates the interaction of global information while preserving local fine-grained information. Furthermore, a bi-directional structure alignment is designed to maximize the inter-modality complementarity.}
  \label{fig:fusion strategy}
\end{figure}

Previous visual-LiDAR odometry works can be classified into two categories: traditional methods \cite{wang2021dv, yuan2023sdv, huang2020lidar, graeter2018limo} and learning-based methods \cite{valente2019deep, zhuoins20234drvo, sun2023transfusionodom}. Traditional methods accomplish the odometry task through a pipeline consisting of feature extraction, frame-to-frame feature matching, motion estimation, and optimization \cite{yuan2023sdv}. However, these methods suffer from inaccurate pose estimation because of the poor quality and low resolution of extracted features \cite{wang2021dv}. With the development of deep learning, some methods \cite{valente2019deep, huang2020epnet} attempt to utilize CNN-based methods for visual-LiDAR fusion and pose estimation. However, the receptive field of feature fusion is limited by the stride and kernel size of CNN. To enlarge the receptive field for multi-modal fusion, attention-based methods are recently proposed, which leverage the cross-attention mechanism for the multi-modal fusion \cite{zhuoins20234drvo, sun2023transfusionodom}. Attention-based methods can fuse multi-modal features globally and establish the cross-frame association with larger receptive fields because of their long-range dependencies. However, due to the quadratic computational complexity, attention-based methods commonly require larger computational consumption and longer inference time \cite{liu2021swin}, which challenges the real-time applications \cite{sun2023transfusionodom}. Moreover, previous learning-based methods mostly adopt the only feature-level fusion strategy as illustrated in \cref{fig:fusion strategy} a), which fails to capture fine-grained pixel-to-point correspondences \cite{liu2023learning}. Recently, some networks \cite{li2023logonet,zhuoins20234drvo} design point-to-image projection and local feature aggregation as in \cref{fig:fusion strategy} b). However, their performances are still limited by the intrinsic data structure misalignment between sparse LiDAR points and dense camera pixels \cite{liu2023learning}.


To address these problems, we propose a novel local-to-global fusion network (DVLO) with bi-directional structure alignment in \cref{fig:fusion strategy} c). Our fusion module consists of two parts: 1) An image is first viewed as a set of pseudo points inspired by \cite{ma2023image} for fine-grained local fusion with LiDAR points (\textit{image-to-point structure alignment}). 2) Point clouds are also converted into pseudo images by cylindrical projection for global adaptive fusion (\textit{point-to-image structure alignment}). Specifically, a novel clustering-based local fusion module (Local Fuser) is designed to perform local fine-grained feature fusion. We first project LiDAR points onto the image plane based on their coordinate calibration matrices to find the corresponding pixels as cluster centers. Meanwhile, for the image-to-point structure alignment, image pixels are reshaped as a set of pseudo points \cite{ma2023image}. Then, within a certain scope of each cluster center, we aggregate pseudo point features to generate the local fused features based on the similarities to the cluster centers dynamically. In the global fusion module, we project point clouds onto the cylindrical surface to obtain pseudo images. Then, an adaptive fusion mechanism is leveraged to merge the above local fused image features and point (pseudo image) features for global fusion. It is worth noting that our fusion module is hierarchically utilized at multi-scale feature maps between images and points. The local fusion module can provide more fine-grained point-to-pixel correspondence information, while the global fusion has a larger receptive field and achieves more global information interaction.

Overall, our contributions are as follows:
\begin{itemize}
	\item[1.] We propose a local-to-global fusion odometry network with bi-directional structure alignment. We cluster image pixels viewed as a set of pseudo points for local fusion with LiDAR points. Point clouds are also converted into pseudo images through cylindrical projection for global adaptive fusion.  
	\item[2.] A pure clustering-based fusion module is designed to obtain the fine-grained local fused features. To the best of our knowledge, our method is the first deep clustering-based multi-modal fusion attempt, serving as an effective and efficient fusion strategy alternative apart from CNN and transformer.
	\item[3.] Extensive experiments on the KITTI odometry dataset \cite{geiger2012we, geiger2013vision} demonstrate that our method outperforms all recent deep LiDAR, visual, and visual-LiDAR fusion odometry works on most sequences. Furthermore, our fusion strategy can generalize well to other multi-modal tasks, like scene flow estimation, even surpassing recent SOTA method CamLiRAFT \cite{liu2023learning}.
\end{itemize}

\section{Related Work}
\label{sec:related}
\noindent\textbf{Deep Visual Odometry.}
Recently, learning-based methods have shown impressive performance in the visual odometry field \cite{konda2015learning,kendall2015posenet,wang2017deepvo,naumann2023nerf,shan2024enerf}. The pioneering work \cite{konda2015learning} uses deep neural networks for odometry estimation with the prediction of both speed and direction for individual images. PoseNet \cite{kendall2015posenet} initially employs Convolution Neural Networks (CNNs) to extract features from the input image and then estimate the pose. DeepVO \cite{wang2017deepvo} applies deep recurrent neural networks to capture the temporal dynamics and interdependency information of sequences, thereby facilitating the estimation of ego-motion. Li \etal \cite{li2022cross} utilize the knowledge distillation technique based on pre-trained visual-LiDAR odometry as a teacher for guiding the training of the visual odometry. Deng \etal \cite{deng2023long} propose a long-term visual SLAM system with map prediction and dynamics removal. NeRF-VO \cite{naumann2023nerf} improves the geometric accuracy of the scene representation by optimizing a set of keyframe poses and the underlying dense geometry through training the radiance field with volume rendering \cite{wang2023animatabledreamer,wang2024vidu4d}.

\noindent\textbf{Deep LiDAR Odometry.} In contrast to visual odometry, deep LiDAR odometry is still a challenging task because of the large number, irregularity, and sparsity of raw LiDAR points \cite{wang2022efficient,chen2023sira,zhang2024comprehensive}. Nicolai \etal \cite{nicolai2016deep} first introduce the deep learning technique into LiDAR odometry. They project 3D LiDAR points onto the 2D plane to obtain 2D depth images, and then employ 2D learning methods for pose estimation. DeepPCO \cite{wang2019deeppco} projects point clouds to panoramic depth images and applies two sub-networks to estimate the translation and rotation respectively. LO-Net \cite{li2019net} also converts points to 2D format by projection and uses the normal of each 3D point and dynamical masks to further improve the performance. PWCLO \cite{wang2021pwclo} introduces the PWC \cite{sun2018pwc} structure for the LiDAR odometry task, which hierarchically refines estimated poses by an iterative warp-refinement module. EfficientLO \cite{wang2022efficient} proposes a projection-aware operator for improving the efficiency of LiDAR odometry. TransLO \cite{liu2023translo} designs a window-based masked point transformer to enhance global feature embeddings and remove outliers. DELO \cite{ali2023delo} introduces partial optimal transportation of LiDAR descriptors and predictive uncertainty for robust pose estimation. NeRF-LOAM \cite{deng2023nerf} applies a neural radiation field to the LiDAR odometry system, showing excellent generalization capabilities across various environments. 

\noindent\textbf{Viusal-LiDAR Odometry.} Recently, there has been an increasing focus on visual-LiDAR odometry, which takes advantage of both 2D texture and 3D geometric features. Existing visual-LiDAR odometry can be classified into two categories: traditional methods and learning-based methods. For traditional methods, V-LOAM \cite{zhang2015visual} leverages the high-frequency estimated poses from visual odometry as a motion prior for the low-frequency LiDAR odometry, resulting in refined motion estimation. LIMO \cite{graeter2018limo} utilizes depth information derived from LiDAR points to alleviate the scale uncertainty that is intrinsic to monocular visual odometry. PL-LOAM \cite{huang2020lidar} provides a pure visual motion tracking method and a novel scale correction algorithm. DV-LOAM \cite{wang2021dv} is a SLAM framework including a two-stage direct visual odometry module, a LiDAR mapping module with considerations of dynamic objects, and a parallel global and local search loop closure detection module. SDV-LOAM \cite{yuan2023sdv} combines the semi-direct visual odometry with an adaptive sweep-to-map LiDAR odometry to tackle the challenges of 3D-2D depth correlation. For learning-based methods, MVL-SLAM \cite{an2022visual} employs the RCNN network architecture, fusing RGB images and multi-channel depth images from 3D LiDAR points. LIP-Loc \cite{shubodh2024lip} proposes a pre-training strategy for cross-modal localization, which utilizes contrastive learning to jointly train image and point encoders. However, the problem of natural data structure inconsistency between points and images has not been fully considered before. To the best of our knowledge, our work is the first visual-LiDAR odometry network with bi-directional structure alignment.

\section{Methodology}
\subsection{Overall Architecture}
\label{sec:Overall Architecture}

\begin{figure}[tb]
  \centering
  \includegraphics[width=1.0\linewidth]{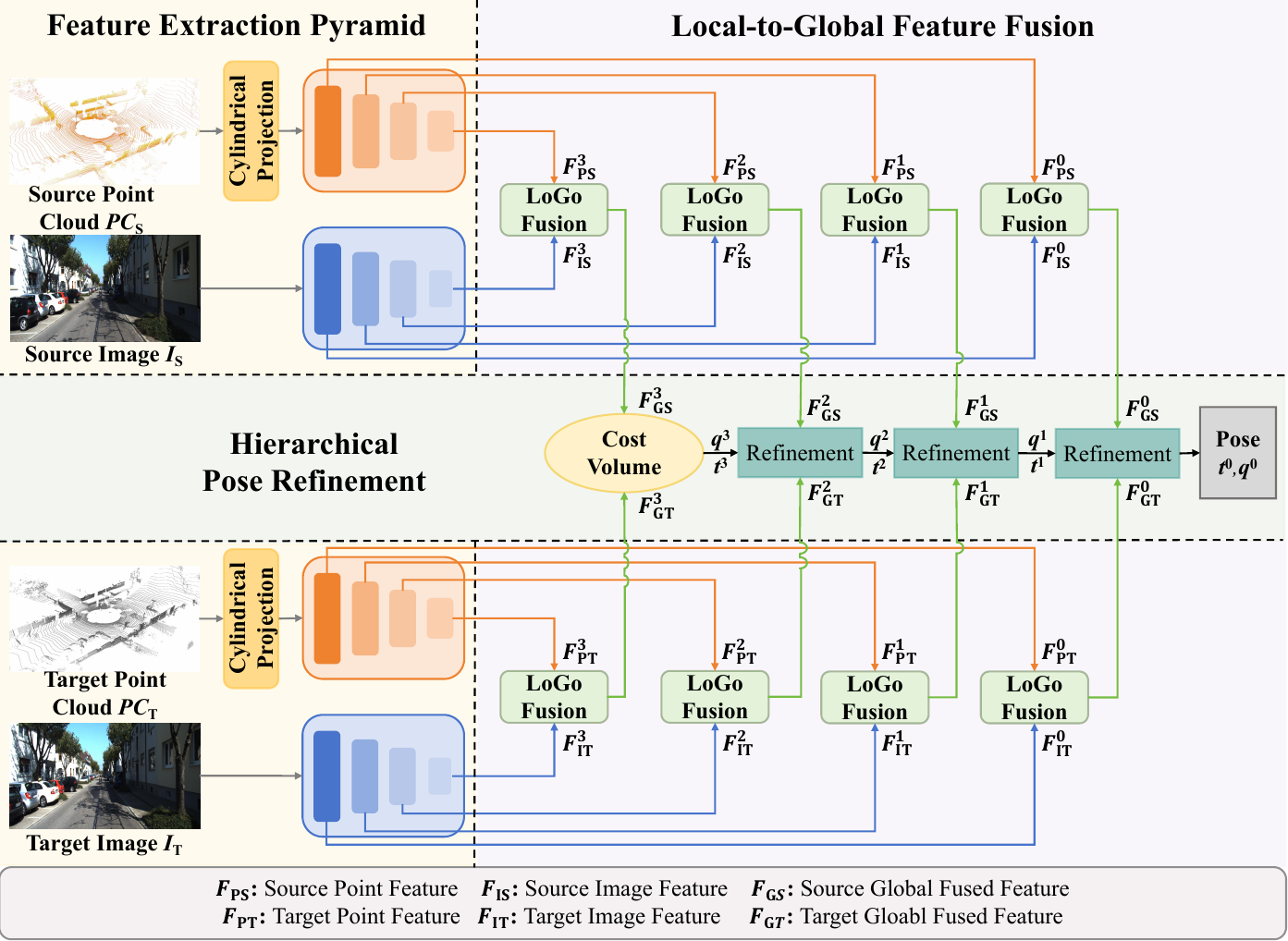}
  \caption{The pipeline of our proposed DVLO. We propose a novel Local-to-Global (LoGo) fusion module, which consists of a clustering-based Local Fuser and an adaptive Global Fuser. The pose is initially regressed from the cost volume of the coarsest fused features and then refined iteratively from fused features in shallower layers.}
  \label{fig:Overview}
\end{figure}

The overall architecture of our proposed DVLO is illustrated in \cref{fig:Overview}. Given two point clouds $PC_{S}$, $PC_{T}  \in \mathbb{R}^{N \times 3} $ and their corresponding monocular camera images $I_{S}$, $I_{T} \in \mathbb{R}^{H \times W \times 3 }$ from a pair of consecutive frames, the goal of our odometry is to estimate the relative pose including both rotation quaternion $q \in \mathbb{R}^{4}$ and translation vector $t \in \mathbb{R}^3$ between two frames. 


In detail, we first feed the camera images and LiDAR points into the hierarchical feature extraction module in \cref{sec:PseudoImage} to obtain multi-level image features and point features. Then, in \cref{sec:LocalFuser} and \cref{sec:GlobalFuser}, we delve into the details of our designed local-to-global fusion module. Finally, poses are estimated and iteratively refined from fused features as in \cref{sec:refinement}.

\subsection{Hierarchical Feature Extraction}
\label{sec:PseudoImage}
\noindent\textbf{Point Feature Extraction.}
Due to the irregularity and sparsity of the raw point clouds, we first project them onto a cylindrical surface \cite{wang2022efficient,liu2023translo} to orderly organize points. Their corresponding 2D positions are:
  \begin{gather}
    u = \arctan 2\left ( y/x \right )/\Delta  \theta, \\
    v = \arcsin\left ( z/\sqrt{x^{2} + y^{2} + z^{2}}  \right ) /\Delta \phi, 
  \end{gather}
where $x$, $y$, $z$ are the raw 3D coordinates of point cloud. $u$, $v$ are the corresponding 2D pixel positions on the projected pseudo image. $\Delta \theta$ and $\Delta \phi$ are horizontal and vertical resolutions of the LiDAR sensor, respectively. To make the best use of geometric information of raw 3D points, we fill each projected 2D position with its corresponding original 3D coordinates. In this case, LiDAR points can not only be converted into the pseudo image structure \cite{li2019net} for better alignment and global feature fusion with images in \cref{sec:GlobalFuser}, but also retain the original 3D geometric information for effective feature extraction. Then, pseudo images of size $H_{P} \times W_{P} \times 3 $ in \cref{fig:Overview} will be fed into the hierarchical feature extraction module \cite{wang2022efficient} to extract multi-level point features $F_{P}\in \mathbb{R}^{H_{P} \times W_{P} \times D}$, where $D$ is the number of channels of the pseudo image features.

\noindent\textbf{Image Feature Extraction.} Given the camera images $I \in \mathbb{R}^{H \times W \times 3}$, we utilize the convolution-based feature pyramid in \cite{huang2020epnet} to extract image features $F_{I} \in \mathbb{R}^{H_{I} \times W_{I} \times C}$, where $H_{I}$, $W_{I}$ are the height and width of the feature map. $C$ is the number of channels of the image features.

\subsection{Local Fuser Module}
\label{sec:LocalFuser}

\begin{figure}[t]
  \centering
  \includegraphics[width=1.0\linewidth]{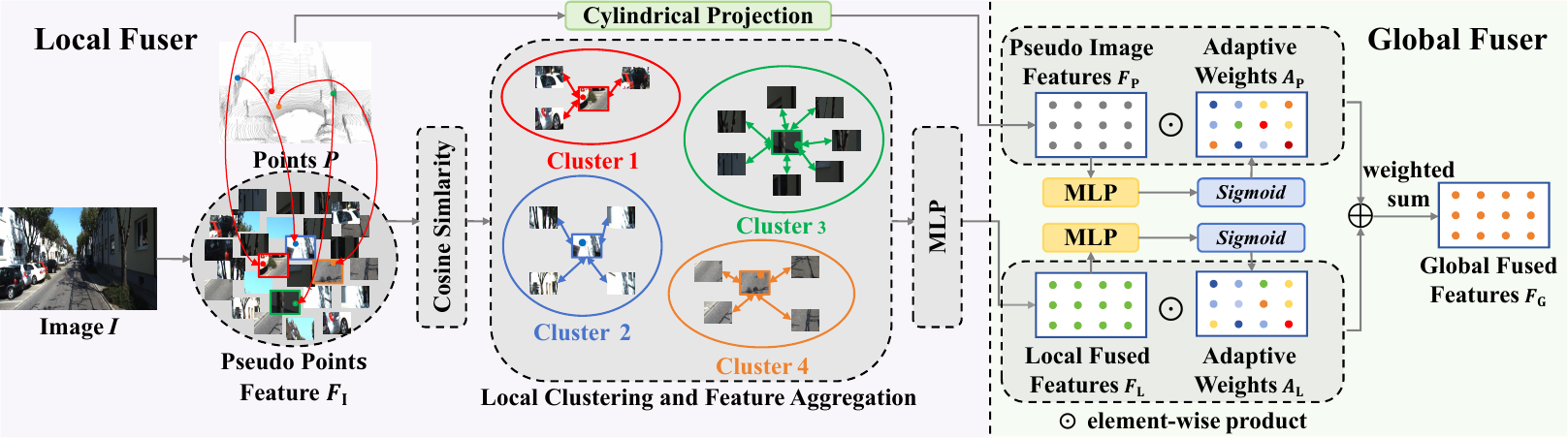}
  \caption{Our designed Local-to-Global (LoGo) Fusion module. We project points onto the image plane based on the coordinate system transformation matrix as cluster centers and convert the image into a set of pseudo points. Then, we locally aggregate pseudo point features based on the similarities to each cluster center.}
  \label{fig:localfuser}
\end{figure}

Inspired by Context Clusters \cite{ma2023image} which proposes a generic clustering-based visual backbone viewing images as a set of points, we extend it and propose a novel clustering-based feature fusion module (Local Fuser) without any CNN or transformer. The module can locally merge more fine-grained 2D texture from images and geometric features from points within each cluster as shown in \cref{fig:localfuser}. Our clustering-based method also maintains high efficiency, where the total inference time is only half of the attention-based methods as in \cref{tab:local stratrgy}.

\noindent\textbf{From Image to Pseudo Points.} Given image features $F_{I} \in \mathbb{R}^{H_{I} \times W_{I} \times C}$ , we first reshape them as a collection of pseudo points $F_{pp} \in \mathbb{R}^{M \times C}$, where $M = H_{I} \times W_{I}$ is the number of pseudo points. In this case, images have the same data structure as LiDAR points, which facilitates the local pixel-to-point correspondence establishment and further clustering-based feature aggregation.

\noindent\textbf{Pseudo Point Clustering.} We first project LiDAR points onto the image plane to obtain their corresponding 2D coordinates $x'$ and $y'$ in the image coordinate system as cluster centers. The center feature $F_{c} \in \mathbb{R}^{N \times C}$ is computed by the bilinear interpolation on $F_{I}$ based on $x'$, $y'$. Then, we divide all pseudo points into several clusters according to the pair-wise cosine similarities between center features $F_{c}$ and pseudo point features $F_{pp}$. Here, we allocate each pseudo point to the most similar center, resulting in $N$ clusters. For efficiency, following Swin Transformer \cite{liu2021swin}, we use the region partition while computing similarity. 

\noindent\textbf{Local Feature Aggregation.} Following \cite{ma2023image}, we aggregate all pseudo point features within the same cluster based on the similarities to the cluster center dynamically. Given the cluster which contains $k$ pseudo points around the $i$-th cluster center, the local fused feature $F_{L}^{i} \in \mathbb{R}^{1 \times C}$ is calculated by:
  \begin{gather}
    F_{L}^{i} = \frac{1}{X} \left ( F_{c}^{i} + \sum_{j=1}^{k} sigmoid\left ( \alpha s_{ij} +\beta \right ) \cdot F_{pp}^{j}  \right ),
    \label{eq:aggregating} \\
    X = 1 + \sum_{j=1}^{k} sigmoid\left ( \alpha s_{ij} +\beta \right ),  
  \end{gather}
where $F_{pp}^{j}$ is the j-th pseudo point's feature. $s_{ij}$ is the similarity score between the $j$-th pseudo point and $i$-th cluster center. $\alpha$ and $\beta$ are learnable scalars to scale and shift the similarity. $sigmoid\left( \cdot \right)$ is a sigmoid function to re-scale the similarity to (0, 1). $X$ is the normalization factor. Since we project LiDAR points onto the image plane as cluster centers and aggregate features for each center, local fused feature $F_{L} \in \mathbb{R}^{N \times C}$ has the same dimension with original LiDAR points. Therefore, we can also reshape the local fused feature $F_{L}$ like a pseudo image with the size of $H_{P} \times W_{P} \times C$ as the input of the Global Fuser module.


\subsection{Global Fuser Module}
\label{sec:GlobalFuser} 
Since local feature fusion is only conducted within a partitioned region, the above Local Fuser module has a limited receptive field. To expand the receptive field for sufficient feature fusion, we introduce the global adaptive fusion mechanism \cite{peng2023delflow} between local fused feature $F_{L}$ and point (pseudo image) feature $F_{P}$ as shown in \cref{fig:localfuser}. Our Global Fuser can enable the global information interaction between two modalities, which facilitates to recognize dynamics and occlusion because they introduce inconsistent global motions \cite{Liu_2023_ICCV}. 

\noindent\textbf{From Points to Pseudo Image.} We convert the sparse LiDAR points to structured pseudo images by the cylindrical projection in \cref{sec:PseudoImage}. In this case, point feature $F_{P}$ has the size of ${H_{P} \times W_{P} \times D}$. This process reorganizes the originally sparse unstructured points as dense structured pseudo images, enabling the following dense feature map fusion with image features.

\noindent\textbf{Adaptive Fusion.} Given the local fused features $F_{L} \in \mathbb{R}^{H_{P} \times W_{P} \times C}$ and point features $F_{P} \in \mathbb{R}^{H_{P} \times W_{P} \times D}$, we perform the adaptive global fusion as:
\begin{gather}
  A_{P} = sigmoid(MLP(F_{P})), \\
  A_{L} = sigmoid(MLP(F_{L})), \\
  F_{G} = \frac{A_{P}\odot F_{P} + A_{L}\odot F_{L}}{A_{p}+A_{L}}, 
\end{gather}
where $A_{P}$ and $A_{L}$ are the adaptive weights for point (pseudo image) features and local fused features, which are obtained by the sigmoid function and MLP layers. $\odot$ represents the element-wise product. We then reshape the global fused feature $F_{G}$ back to the size of $ N \times D $ as the input of the iterative pose estimation.

\subsection{Iterative Pose Estimation}
\label{sec:refinement} 
Following \cite{wang2022efficient, wang2021pwclo}, we use the attentive cost volume to generate the coarse embedding features $E \in \mathbb{R}^{N \times D}$ by associating global fused features $F_{GS}^{3}$ and $F_{GT}^{3}$ of two frames in the coarsest layer. The embedding features contain the correlation information between two consecutive frames. Then, we utilize the weighting embedding mask $M$ on the embedding features $E$ to regress the pose transformation. The weighting embedding mask $M$ is calculated by:
  \begin{gather}
    M = softmax(MLP(E \oplus F_{GS}^{3})),
  \end{gather}
where $M \in \mathbb{R}^{N \times D}$ is the learnable masks. $F_{GS}^{3} \in \mathbb{R}^{N \times D}$ is the global fused features in the source frame. Then, the quaternion $q \in \mathbb{R}^{4}$ and translation vector $t \in \mathbb{R}^3$ are generated by weighting embedding features and FC layers:
  \begin{gather}
    q = \frac{FC(E \odot M)}{\left | FC(E \odot M) \right | },
    \label{eq:q} \\
    t = FC(E \odot M).
    \label{eq:t}
  \end{gather}
After the initial estimation $q$ and $t$, we refine them by the iterative refinement module in \cite{wang2021pwclo} to get the final pose. The refined quaternion $q^{l}$ and translation vector $t^{l}$ of the $l$ layer can be calculated by:
  \begin{gather}
    q^{l} = \Delta q^{l}q^{l+1},  \\
    [0, t^{l}] = \Delta q^{l}[0, t^{l+1}](\Delta q^{l})^{-1} + [0, \Delta t^{l}], 
  \end{gather}
where pose residuals $\Delta q^{l}$ and $\Delta t^{l}$ can be obtained by the similar process in the coarsest layer following \cref{eq:q} and \cref{eq:t}.

\subsection{Loss Function}
Our network outputs $q^{l}$ and $t^{l}$ from four layers will be involved to calculate the supervised loss $\mathcal{L}^{l}$ \cite{wang2022efficient, wang2021pwclo}. The training loss function of $l$-th layer is:
  \begin{gather}
    \mathcal{L}^{l} = \left \| t_{gt} - t^{l} \right \| exp(-k_{x}) + k_{x} + \left \|q_{gt}-{q^{l}}   \right \|_{2}exp(-k_{q}) +k_{q},
  \end{gather}
where $t_{gt}$ and $q_{gt}$ are the ground truth translation and quaternion, respectively. $k_{x}$ and $k_{q}$ are the learnable scalars to scale the loss. $\left \| \cdot \right \|$ and $\left \| \cdot \right \|_{2}$ are the $L_{1}$ and $L_{2}$ norm, respectively. Then, the total training loss is:
  \begin{gather}
    \mathcal{L} = \sum_{l=1}^{L} \alpha ^{l}\mathcal{L}^{l},
  \end{gather} 
where L is the total number of layers (set as 4), and $\alpha^{l}$ is a hyperparameter representing the weight of $l$ layer.

\begin{table}[t]
  \centering
  \caption{Comparison with different odometry networks on the KITTI odometry dataset \cite{geiger2013vision}. $t_{rel}$ and $r_{rel}$ mean the average sequence translational RMSE (\%) and the average sequence rotational RMSE ($^{\circ}$/100m) respectively on 00-10 subsequences in the length of 100, 200, ..., 800m. The best results are {\color{red}\bf{bold}}, and the second best results are {\color{blue}\underline{underlined}}. $^{*}$ represents the model is trained on the 00-08 sequences.}
  \resizebox{1\columnwidth}{!}{
    \begin{tabular}{lllll|cc|cc|cc|cc|cc|cc|cc|cc|cc|cc|cc|cc}
    \toprule
    \multicolumn{5}{l|}{\multirow{2}[0]{*}{\textbf{Method}}} & \multicolumn{2}{c|}{00} & \multicolumn{2}{c|}{01} & \multicolumn{2}{c|}{02} & \multicolumn{2}{c|}{03} & \multicolumn{2}{c|}{04} & \multicolumn{2}{c|}{05} & \multicolumn{2}{c|}{06} & \multicolumn{2}{c|}{07} & \multicolumn{2}{c|}{08} & \multicolumn{2}{c|}{09} & \multicolumn{2}{c|}{10} & \multicolumn{2}{c}{mean (07-10)} \\
         &      &      &      &      & $t_{rel}$ & $r_{rel}$ & $t_{rel}$ & $r_{rel}$ & $t_{rel}$ & $r_{rel}$ & $t_{rel}$ & $r_{rel}$ & $t_{rel}$ & $r_{rel}$ & $t_{rel}$ & $r_{rel}$ & $t_{rel}$ & $r_{rel}$ & $t_{rel}$ & $r_{rel}$ & $t_{rel}$ & $r_{rel}$ & $t_{rel}$ & $r_{rel}$ & $t_{rel}$ & $r_{rel}$ & $t_{rel}$ & $r_{rel}$\\
    \midrule
    \multicolumn{29}{l}{\it{Visual Odometry Methods:}} \\
    \midrule
    \multicolumn{5}{l|}{SfMLearner$^{*}$\cite{zhou2017unsupervised}}         &21.32&6.19&22.41&2.79&24.10&4.18&12.56&4.52&4.32&3.28&12.99&4.66&15.55&5.58&12.61&6.31&10.66&3.75&11.32&4.07&15.25&4.06&12.46&4.55\\
    \multicolumn{5}{l|}{MLM-SFM$^{*}$\cite{song2015high}}  &2.04&0.48&--&--&1.50&0.35&3.37&{\color{red}\bf{0.21}}&1.43&0.23&2.19&0.38&2.09&0.81&--&--&2.37&{\color{blue}\underline{0.44}}&1.76&0.47&2.12&0.85&2.08&0.59\\
    \multicolumn{5}{l|}{DFVO$^{*}$\cite{zhan2021df}}                &2.01&0.79&61.17&18.96&2.46&0.79&3.27&0.89&0.79&0.56&1.50&0.74&1.95&0.76&2.28&1.16&2.11&0.74&3.21&0.59&2.89&0.97&2.62&0.87\\
    \multicolumn{5}{l|}{Cho \etal$^{*}$ \cite{cho2023dynamic}}      &1.77&0.79&64.38&16.87&2.62&0.74&3.06&0.89&0.65&0.55&1.31&0.74&1.60&0.56&1.06&0.67&2.28&0.76&2.66&0.53&2.95&0.95&2.24&0.73\\
    \midrule
    \multicolumn{29}{l}{\it{LiDAR Odometry Methods:}} \\
    \midrule
    \multicolumn{5}{l|}{LO-Net \cite{li2019net}}           &1.47&0.72&1.36&0.47&1.52&0.71&1.03&0.66&0.51&0.65&1.04&0.69&0.71&0.50&1.70&0.89&2.12&0.77&1.37&0.58&1.80&0.93&1.75&0.79\\
    \multicolumn{5}{l|}{PWCLO\cite{wang2021pwclo}}           &0.89&0.43&1.11&0.42&1.87&0.76&1.42&0.92&1.15&0.94&1.34&0.71&0.60&0.38&1.16&1.00&1.68&0.72&0.88&0.46&2.14&0.71&1.47&0.72\\
    \multicolumn{5}{l|}{DELO\cite{ali2023delo}}              &1.43&0.81&2.19&0.57&1.48&0.52&1.38&1.10&2.45&1.70&1.27&0.64&0.83&0.35&0.58&0.41&1.36&0.64&1.23&0.57&1.53&0.90&1.18&0.63\\
    \multicolumn{5}{l|}{TransLO\cite{liu2023translo}}        &0.85&0.38&1.16&0.45&{\color{blue}\underline{0.88}}&0.34&1.00&0.71&{\color{blue}\underline{0.34}}&{\color{blue}\underline{0.18}}&0.63&0.41&0.73&0.31&0.55&0.43&1.29&0.50&0.95&0.46&1.18&0.61&0.99&0.50\\
    \multicolumn{5}{l|}{EfficientLO\cite{wang2022efficient}} &{\color{red}\bf{0.80}}&{\color{blue}\underline{0.37}}&{\color{blue}\underline{0.91}}&{\color{blue}\underline{0.40}}&0.94&{\color{blue}\underline{0.32}}&{\color{red}\bf{0.51}}&0.43&0.38&0.30&{\color{blue}\underline{0.57}}&{\color{blue}\underline{0.33}}&{\color{blue}\underline{0.36}}&{\color{blue}\underline{0.23}}&{\color{red}\bf{0.37}}&{\color{red}\bf{0.26}}&{\color{blue}\underline{1.22}}&0.48&{\color{blue}\underline{0.87}}&0.38&{\color{blue}\underline{0.91}}&{\color{blue}\underline{0.50}}&{\color{blue}\underline{0.86}}&{\color{red}\bf{0.41}}\\
    \midrule
    \multicolumn{29}{l}{\it{Multimodal Odometry Methods:}} \\
    \midrule
    \multicolumn{5}{l|}{An \etal$^{*}$  \cite{an2022visual}}  &2.53&0.79&3.76&0.80&3.95&1.05&2.75&1.39&1.81&1.48&3.49&0.79&1.84&0.83&3.27&1.51&2.75&1.61&3.70&1.83&4.65&0.51&3.59&1.37\\
    \multicolumn{5}{l|}{H-VLO$^{*}$\cite{aydemir2022h}}             &1.75&0.62&43.2&0.46&2.32&0.60&2.52&0.47&0.73&0.36&0.85&0.35&0.75&0.30&0.79&0.48&1.35&{\color{red}\bf{0.38}}&1.89&{\color{red}\bf{0.34}}&1.39&0.52&1.36&{\color{blue}\underline{0.43}}\\
    \multicolumn{5}{l|}{Ours}                                 &{\color{red}\bf{0.80}}&{\color{red}\bf{0.35}}&{\color{red}\bf{0.85}}&{\color{red}\bf{0.33}}&{\color{red}\bf{0.81}}&{\color{red}\bf{0.29}}&{\color{blue}\underline{0.59}}&{\color{blue}\underline{0.36}}&{\color{red}\bf{0.26}}&{\color{red}\bf{0.13}}&{\color{red}\bf{0.41}}&{\color{red}\bf{0.23}}&{\color{red}\bf{0.33}}&{\color{red}\bf{0.17}}&{\color{blue}\underline{0.46}}&{\color{blue}\underline{0.33}}&{\color{red}\bf{1.09}}&{\color{blue}\underline{0.44}}&{\color{red}\bf{0.85}}&{\color{blue}\underline{0.36}}&{\color{red}\bf{0.88}}&{\color{red}\bf{0.46}}&{\color{red}\bf{0.82}}&{\color{red}\bf{0.41}}\\
    \bottomrule
    \end{tabular}%
  }
  \label{tab:evaluation}%
\end{table}%

\section{Experiment}
\label{sec:Exper}
\subsection{KITTI Odometry Dataset}
We evaluate our DVLO on the KITTI odometry dataset \cite{geiger2013vision}, which is a widely used benchmark for the evaluation of odometry and SLAM system. The dataset consists of 22 sequences of LiDAR point clouds and their corresponding stereo images. In this paper, we only use the monocular left camera image for the fusion with the LiDAR sensor. Since the ground truth pose (trajectory) is only available for sequences 00-10, we utilize these sequences for training and testing.  

\subsection{Implementation Details}
\label{sec:Implement}
\noindent\textbf{Data Preprocessing.} We directly input all LiDAR points without downsampling. The projected pseudo image size is set in line with the range of the LiDAR sensor as 64 $\times$ 1800. We pad the camera images to a uniform size of 384 $\times$ 1280. Since there is a large spatial range difference between the camera and LiDAR, we design a fusion mask to indicate which point can be fused with the image. 

\noindent\textbf{Parameters.} Experiments are conducted on an NVIDIA RTX 4090 GPU with PyTorch 1.10.1. We use Adam optimizer with $\beta_{1} = 0.9$, $\beta_{2} = 0.999$. The initial learning rate is set to 0.001 and exponentially decays every 200000 steps until 0.00001. Batch size is 8. $\alpha^{l}$ for four layers are 1.6, 0.8, 0.4, and 0.2. Initial values of learnable parameters $k_{x}$ and $k_{q}$ are set as 0.0 and -2.5, respectively.

\noindent\textbf{Evaluation Metrics.} We follow protocols of PWCLO \cite{wang2021pwclo} to evaluate our method with two metrics: (1) Average sequence translational RMSE (\%). (2) Average sequence rotational RMSE ($^{\circ}$/100m).

\setlength{\tabcolsep}{3mm}
\begin{table}[t]
  \centering
  \caption{Comparison with traditional visual-LiDAR odometry on KITTI 00-10 sequences. Our DVLO is trained on 00-06 sequences. The best results for each sequence are {\color{red}\bf{bold}}, and the second best results are {\color{blue}\underline{underlined}}.}
    \resizebox{1.0\textwidth}{!}{
    \begin{tabular}{ll|c|c|c|c|c|c|c|c|c|c|c|c}
    \toprule
    \multicolumn{2}{l|}{\multirow{2}[0]{*}{\textbf{  Method}}} & 00 & 01 & 02 & 03 & 04 & 05 & 06 & 07 & 08 & 09 & 10 & Mean(00-10) \\
    \multicolumn{2}{l|}{} & $t_{rel}$ & $t_{rel}$ & $t_{rel}$ & $t_{rel}$ & $t_{rel}$ & $t_{rel}$ & $t_{rel}$ & $t_{rel}$ & $t_{rel}$ & $t_{rel}$ & $t_{rel}$ & $t_{rel}$ \\
    \midrule
     \multicolumn{2}{l|}{  V-LOAM \cite{zhang2015visual}} &--&--&--&--&--&--&--&--&-- & 1.74  & 1.01  & 1.38   \\
    \multicolumn{2}{l|}{  DVL-SLAM \cite{shin2020dvl}} & {\color{blue}\underline{0.93}}  & {\color{blue}\underline{1.47}}  & {\color{blue}\underline{1.11}}  & 0.92  & 0.67  & 0.82  & 0.92  & 1.26  & 1.32  & \color{red}\bf{0.66}  & \color{red}\bf{0.70}  & 0.98  \\
    \multicolumn{2}{l|}{  PL-LOAM \cite{huang2020lidar}} & 0.99  & 1.87  & 1.38  & {\color{blue}\underline{0.65}}  & {\color{blue}\underline{0.42}}  & {\color{blue}\underline{0.72}}  & {\color{blue}\underline{0.61}}  & {\color{blue}\underline{0.56}}  & {\color{blue}\underline{1.27}}  & 1.06  & {\color{blue}\underline{0.83}}  & {\color{blue}\underline{0.94}}  \\
    \multicolumn{2}{l|}{  Ours} & \color{red}\bf{0.80}  & \color{red}\bf{0.85}  & \color{red}\bf{0.81}  & \color{red}\bf{0.59}  & \color{red}\bf{0.26}  & \color{red}\bf{0.41}  & \color{red}\bf{0.33}  & \color{red}\bf{0.46}  & \color{red}\bf{1.09}  & {\color{blue}\underline{0.85}}  & 0.88  & \color{red}\bf{0.67}  \\
    \bottomrule
    \end{tabular}%
    }
  \label{tab:evalution2}%
\end{table}%

\setlength{\tabcolsep}{6mm}
\begin{table}[t]
  \centering
  \caption{Comparison with learning-based multi-modal odometry on KITTI 09-10 sequences. Our DVLO is trained on 00-06 sequences while other models are trained on 00-08 sequences. The best results are {\color{red}\bf{bold}}, and the second best results are {\color{blue}\underline{underlined}}. }
    \resizebox{1.0\textwidth}{!}{
    \begin{tabular}{lllll|lll|cc|cc|cc}
    \toprule
    \multicolumn{5}{l|}{\multirow{2}[0]{*}{\bf{Method}}} & \multicolumn{3}{c|}{\multirow{2}[0]{*}{{Modalities}}} &\multicolumn{2}{c|}{09} & \multicolumn{2}{c|}{10} & \multicolumn{2}{c}{Mean (09-10)} \\
    \multicolumn{5}{l|}{} & \multicolumn{3}{l|}{} & $t_{rel}$ & $r_{rel}$ & $t_{rel}$ & $r_{rel}$ & $t_{rel}$ & $r_{rel}$ \\
    \midrule
    \multicolumn{5}{l|}{  Self-VLO\cite{li2021self}} & \multicolumn{3}{c|}{visual+LiDAR} & 2.58  & 1.13  & 2.67  & 1.28  & 2.62  & 1.21  \\
    \multicolumn{5}{l|}{  H-VLO\cite{aydemir2022h}} & \multicolumn{3}{c|}{visual+LiDAR} & 1.89  & {\color{red}\bf{0.34}}  & {\color{blue}\underline{1.39}} & {\color{blue}\underline{0.52}}& {\color{blue}\underline{1.64}}  & {\color{blue}\underline{0.43}}  \\
    \multicolumn{5}{l|}{  SelfVIO \cite{almalioglu2022selfvio}} & \multicolumn{3}{c|}{visual+inertial} & 1.95  & 1.15  & 1.81  & 1.30  & 1.88  & 1.23  \\
    \multicolumn{5}{l|}{  VIOLearner \cite{shamwell2019unsupervised}} & \multicolumn{3}{c|}{visual+inertial} & {\color{blue}\underline{1.82}}  & 1.08  & 1.74  & 1.38  & 1.78  & 1.23  \\
    \multicolumn{5}{l|}{  Ours} & \multicolumn{3}{c|}{visual+LiDAR} & {\color{red}\bf{0.85}}  & {\color{blue}\underline{0.36}}  & {\color{red}\bf{0.88}}  & {\color{red}\bf{0.46}}  & {\color{red}\bf{0.87}}  & {\color{red}\bf{0.41}}  \\
    \bottomrule
    \end{tabular}
    }%
  \label{tab:09-10results}%
\end{table}%

\setlength{\tabcolsep}{8mm}
\begin{table}[t]
  \centering
  \caption{Average inference time of different multi-modal odometry methods on the sequence 07-10 of KITTI odometry dataset.}
    \resizebox{1.00\columnwidth}{!}
            {
    \begin{tabular}{l|ccccc}
    \toprule
    Method &DV-LOAM \cite{wang2021dv} & PL-LOAM \cite{huang2020lidar}&OKVIS-S \cite{leutenegger2013keyframe} &Shu \etal \cite{shu2022multi} &Ours \\
    \midrule
   Inference Time & 167 ms &200 ms& 143 ms&100ms&98.5 ms\\
    \bottomrule
    \end{tabular}%
    }
  \label{tab:Runtime}%
\end{table}%

\begin{figure}[t]
  \centering
  \includegraphics[width=1\linewidth]{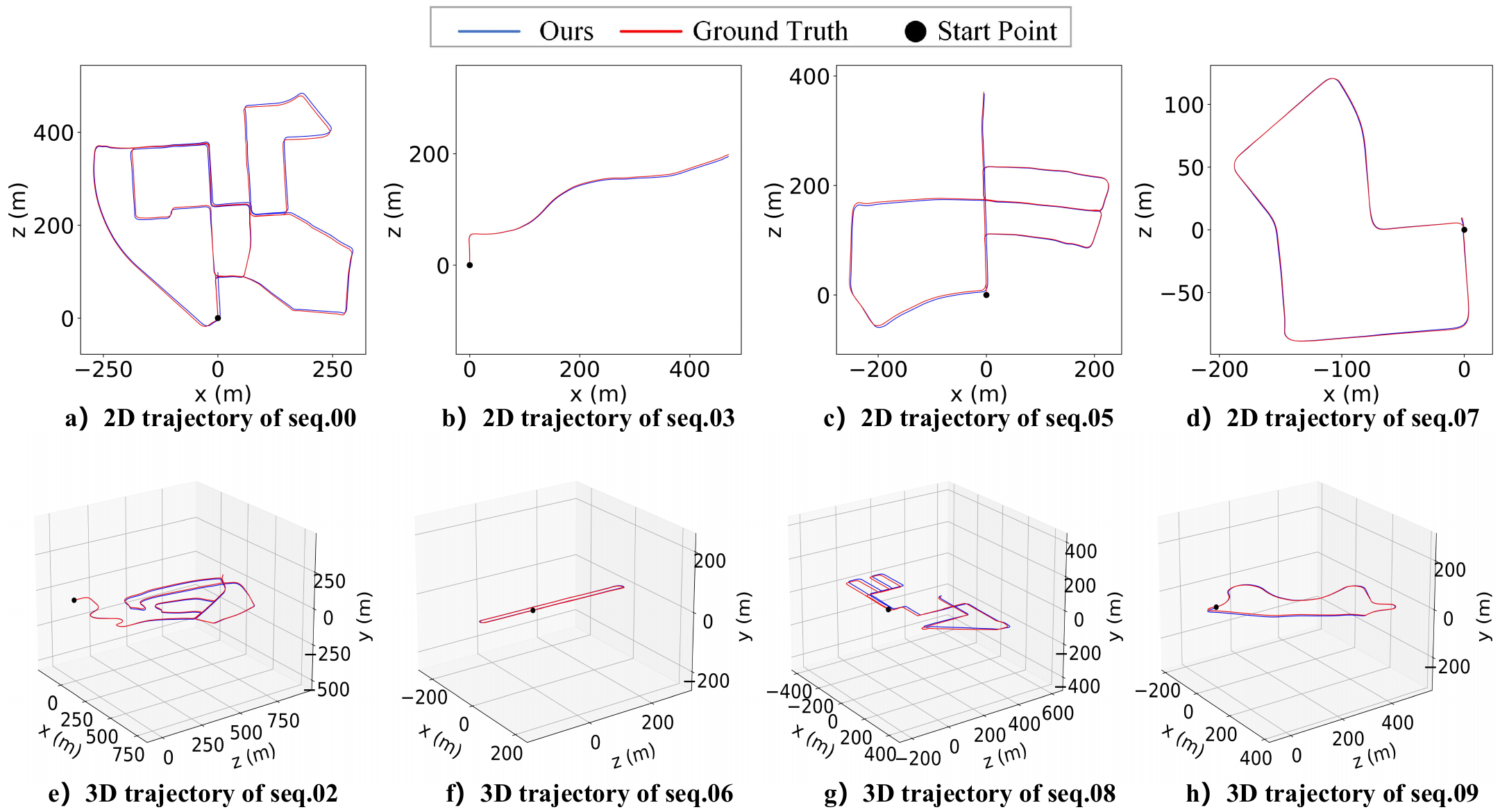}
  \caption{Trajectory of our estimated pose. This figure shows both 2D and 3D trajectories of our network and also the ground truth one on the KITTI dataset.}
  \label{fig:trajectory}
\end{figure}

\begin{figure}[t]
  \centering
  \includegraphics[width=1\linewidth]{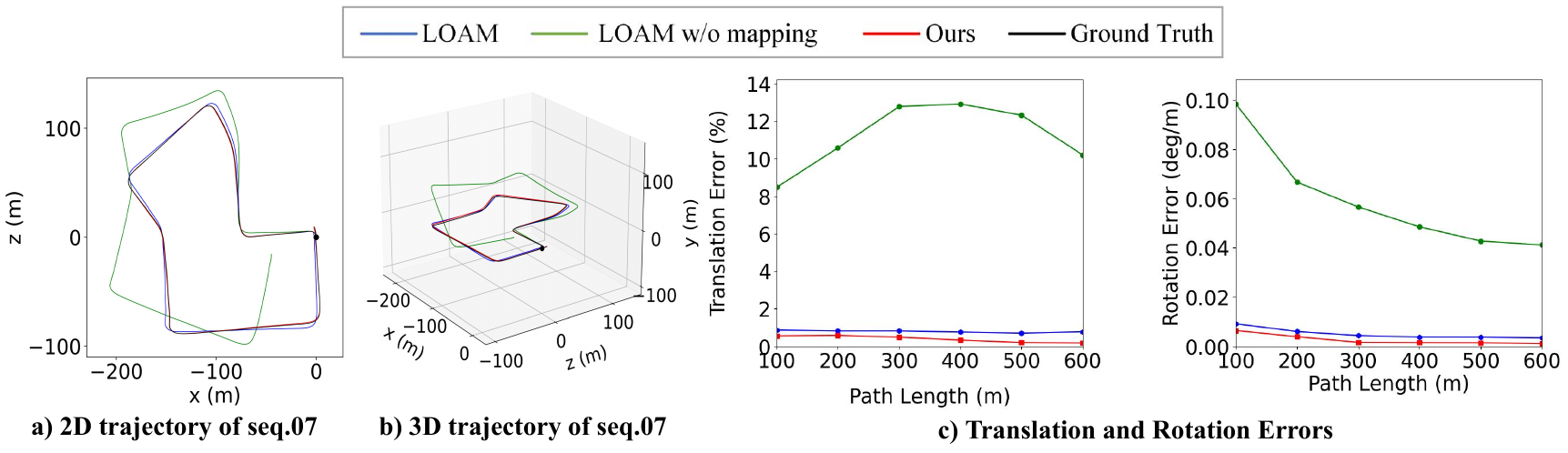}
  \caption{Trajectory results of LOAM and ours on the KITTI sequence 07 with ground truth. Our performance is better than LOAM both without and with mapping.}
  \label{fig:trajectory2}
\end{figure}

\subsection{Quantitative Results}
\noindent\textbf{Comparison with Visual/LiDAR Odometry.} We compare our method with some representative visual odometry (VO) or LiDAR odometry (LO) networks for the comprehensive comparison. Following the settings in \cite{wang2021pwclo}, we train our model on 00-06 sequences. Quantitative results on the KITTI dataset are listed in \cref{tab:evaluation}, which shows that our DVLO outperforms all these works on most sequences. Compared with deep visual odometry, \eg, DFVO\cite{zhan2021df} and Cho \etal.\cite{cho2023dynamic}, our method's mean errors $t_{rel}$ and $r_{rel}$ on sequence 07-10 have a 63.4\% and a 43.8\% decline, respectively. Notably, even though these VO methods are mostly trained on larger data (00-08), our method still outperforms them by a large margin. Compared with deep LiDAR odometry, our DVLO even outperforms the recent SOTA method EfficientLO \cite{wang2022efficient} on most sequences. Compared with EfficientLO, our method has a competitive 0.41 $^{\circ}$/100m rotation error. Moreover, our mean translation error $t_{rel}$ on testing sequences has a 4.9\% decline compared with theirs. The experiment results prove the effectiveness and great potential for our visual-LiDAR fusion design.

\noindent\textbf{Comparison with Traditional Multi-Modal Odometry.} We compare the performance between our method and previous traditional multi-modal odometry works on the whole KITTI sequences (00-10). The results are shown in \cref{tab:evalution2}, which demonstrates that our DVLO outperforms all these works on most sequences. Compared with PL-LOAM \cite{huang2020lidar}, our method's mean translation error $t_{rel}$ on sequence 00-10 has a 28.7\% decline.

\noindent\textbf{Comparison with Learning-based Multi-Modal Odometry.} Because most deep multi-modal fusion odometry methods are trained on the 00-08 sequences and tested on the 09-10 sequences, we also compare the performance between our DVLO and other learning-based multi-modal odometry works on the 09-10 sequences. The results are shown in \cref{tab:evaluation} and \cref{tab:09-10results}. Notably, even though our model is only trained on the 00-06 sequences, our method still outperforms H-VLO on most sequences where our method has a 47.0\% lower $t_{rel}$ and 2.3\% lower $r_{rel}$, which proves the superiority of our proposed fusion strategy.

\begin{figure}[t]
  \centering
  \includegraphics[width=1\linewidth]{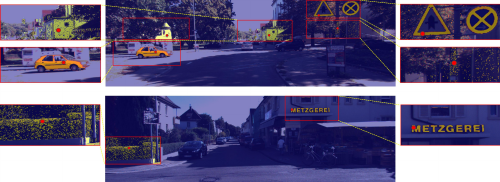}
  \caption{Visualization of our designed local clustering-based fusion mechanism within a certain cluster. Red points indicate the 2D positions of cluster centers. The yellow regions are clustered pixels around each center.}
  \label{fig:clustervisual}
\end{figure}

\subsection{Runtime Analysis}
\label{sec:runtime}
Efficiency is another extremely significant factor in real-time SLAM systems. As shown in \cref{tab:Runtime}, we compare the runtime of our DVLO with other multi-modal odometry methods. Since the LiDAR points in the KITTI dataset are captured at a 10Hz frequency, previous multi-modal methods \cite{wang2021dv,huang2020lidar,leutenegger2013keyframe,shu2022multi} rarely satisfy the real-time application requirements (under 100 ms). However, our method has only 98.5 ms inference time, which has the potential for real-time application. 

\subsection{Visualization Results.}
In this section, we visualize 2D and 3D trajectories based on our estimated pose in \cref{fig:trajectory}. The figure shows that our odometry can well track the trajectory of the ground truth. We also conduct experiments to compare trajectory accuracy and estimation errors between the classical method LOAM \cite{zhang2014loam} and ours. Visualization results are shown in \cref{fig:trajectory2}. Even though our designed odometry is only the front end of the SLAM system without mapping, our method achieves better localization performance than LOAM with mapping.

To illustrate the clustering mechanism in our designed local fusion module, we also visualize the clustered pixels around specific cluster centers. As in \cref{fig:clustervisual}, pixels with similar texture information (yellow regions) are accurately clustered by the point-wise cosine similarity calculation with the cluster centers (red dots).

\setlength{\tabcolsep}{8mm}
\begin{table}[t]
    \centering
      \caption{Comparison with previous scene flow estimation works on the “val” split of the FlyingThings3D subset \cite{mayer2016large}. “RGB” and “XYZ” denote the image and point cloud respectively. The best results are \textbf{bold}.}
 \resizebox{1.00\columnwidth}{!}
            {
 \begin{tabular}{l|c|c|c|c|c|c}
\toprule
\multirow{2}{*}{ \textbf{Method} } & \multirow{2}{*}{ \textbf{Reference} }& \multirow{2}{*}{ \textbf{Input} } & \multicolumn{2}{c}{ \textbf{2D Metrics} } & \multicolumn{2}{c}{ \textbf{3D Metrics} } \\
& & &$\text{EPE}_\text{2D}$ & $\text{ACC}_{1px}$  & $\text{EPE}_\text{3D}$  &  $\text{ACC}_{.05}$  \\
\midrule FlowNet2 \cite{ilg2017flownet} & CVPR'17 & RGB & 5.05 &  72.8 \%  & - & - \\
PWC-Net \cite{sun2018pwc}& CVPR'18 & RGB & 6.55 &  64.3 \%  & - & - \\
RAFT \cite{teed2020raft} & ECCV'20& RGB & 3.12 &  81.1 \%  & - & - \\
\midrule FlowNet3D \cite{liu2019flownet3d}& CVPR'19 & XYZ & - & - & 0.214 &  18.2 \%  \\
PointPWC \cite{wu2020pointpwc}& ECCV'20  & XYZ & - & - &  {0.195}  & - \\
OGSF-Net \cite{ouyang2021occlusion} & CVPR'21  & XYZ & - & - &  {0.163}  & - \\
\midrule 
CamLiFlow \cite{liu2022camliflow}  & CVPR'22& RGB+XYZ & 2.18 &  84.3 \%  & 0.061 &  85.6 \%  \\
DELFlow \cite{peng2023delflow}& ICCV'23 & RGB+XYZ & 2.02 & 85.9 \% & 0.058 & 86.7 \% \\
CamLiRAFT \cite{liu2023learning}& TPAMI'23  & RGB+XYZ &  {1.73}  &  {87.5} \%  &  {0.049}  &  {88.4} \%  \\
\midrule
Ours &---& RGB+XYZ &  \bf{1.69}  &  \bf{87.6} \%  &  \bf{0.048}  &  \bf{88.6} \% \\
\bottomrule
\end{tabular}
}
\label{tab:flow}   
\end{table}

\begin{figure}[t]
  \centering
  \includegraphics[width=1\linewidth]{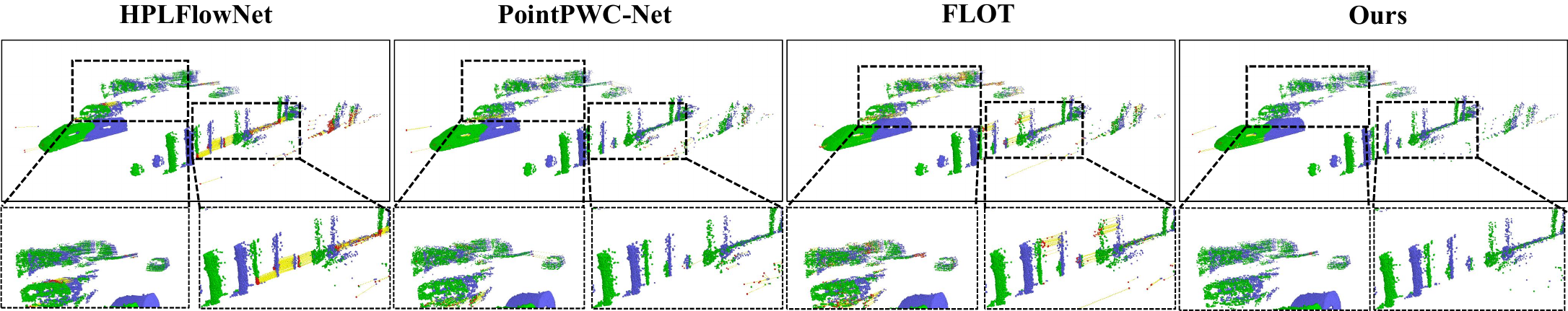}
  \caption{Visualization of our estimated scene flow. Blue points are the source points. Green and red points respectively indicate the correct and wrong estimated target ones.}
  \label{fig:flowvisual}
\end{figure}

\subsection{Generalization to Scene Flow Estimation Task}
It is worth noting that our design can serve as a generic fusion module, which generalizes well to other tasks. Here, we extend our fusion module to the scene flow estimation task \cite{liu2019flownet3d,liu2023learning,liu2024difflow3d,jiang20243dsflabelling,ding2022self,ding2023hidden,zhang2024deflow}. As shown in \cref{tab:flow}, our method surpasses all recent SOTA multi-modal scene flow networks on the FlyingThings3D dataset \cite{mayer2016large} in terms of both 2D and 3D metrics. Our method even consistently outperforms CamLiRAFT \cite{liu2023learning}, which is specially designed for the multi-modal scene flow task. The experiment results demonstrate the strong generalization and universal application capabilities of our method. We also visualize the estimated flow in \cref{fig:flowvisual}.

\subsection{Ablation Study}
In this section, extensive ablation studies are conducted to assess the significance of our designed components.

\noindent\textbf{Without Local Fuser.} We remove the Local Fuser module from our network and directly fuse the image and point features using the Global Fuser module. The results in \cref{tab:ablationl} show that the performance of our model drops significantly without Local Fuser. This demonstrates the importance of the Local Fuser module since local point-to-pixel correspondences can merge more fine-grained features from different modalities.

\noindent\textbf{Without Global Fuser.} We remove the Global Fuser module and directly use the local fused features for pose estimation. Results in \cref{tab:ablationl} demonstrate that the limited receptive field of the Local Fuser module can not enable sufficient global information interaction, which leads to about 13.4\% higher $t_{rel}$ and 14.6\% higher $r_{rel}$. The global modeling ability of Global Fuser can help to recognize outliers that are harmful to pose regression \cite{Liu_2023_ICCV}.

\setlength{\tabcolsep}{5mm}
\begin{table}[t]
  \centering
  \caption{Significance of Local Fuser (LoF) and Global Fuser (GoF) in our local-to-global fusion network. The best results for each sequence are \bf{bold}.}
    \resizebox{1.00\columnwidth}{!}
            {
    \begin{tabular}{cc|cc|cc|cc|cc|cc}
    \toprule
        &   & \multicolumn{2}{c|}{07} & \multicolumn{2}{c|}{08} & \multicolumn{2}{c|}{09} & \multicolumn{2}{c|}{10} & \multicolumn{2}{c}{Mean (07-10)} \\
      \multirow{-2}{*}{\begin{tabular}[c]{@{}c@{}}LoF \end{tabular}} & \multirow{-2}{*}{\begin{tabular}[c]{@{}c@{}}GoF \end{tabular}} & $t_{rel}$ & $r_{rel}$ & $t_{rel}$ & $r_{rel}$ & $t_{rel}$ & $r_{rel}$ & $t_{rel}$ & $r_{rel}$ & $t_{rel}$ & $r_{rel}$ \\
    \midrule
      $\surd$ &   & 0.48  & \bf{0.34}  & 1.11  & 0.50  & 1.11  & 0.50  & 1.02  & 0.54  & 0.93  & 0.47  \\
      & $\surd $  & 0.68  & 0.45  & 1.30  & 0.52  & 1.00  & 0.49  & 1.02  & 0.55  & 1.00  & 0.50  \\
      $\surd $ & $\surd $  & \bf{0.46} &\bf{0.34}  & \bf{1.09} & \bf{0.44}  & \bf{0.85}  & \bf{0.36}  & \bf{0.88}  & \bf{0.48} & \bf{0.82}  & \bf{0.41}  \\
     
    \bottomrule
    \end{tabular}}
  \label{tab:ablationl}%
\end{table}%

\setlength{\tabcolsep}{4mm}
\begin{table}[t]
  \centering
  \caption{Ablation studies of different local fusion strategies. Our clustering-based local fusion strategy can achieve the highest accuracy with half inference time compared with attention-based ones. The best results in accuracy and efficiency are \bf{bold}.}
  \resizebox{1.0\textwidth}{!}{
    \begin{tabular}{ll|cc|cc|cc|cc|cc|cc}
    \toprule
    \multicolumn{2}{l|}{\multirow{2}[0]{*}{\textbf{  Method}}} & \multicolumn{2}{c|}{07} & \multicolumn{2}{c|}{08} & \multicolumn{2}{c|}{09} & \multicolumn{2}{c|}{10} & \multicolumn{2}{c|}{Mean (07-10)} & \multicolumn{2}{c}{\multirow{2}[0]{*}{Inference Time}} \\
    \multicolumn{2}{l|}{} & $t_{rel}$ & $r_{rel}$ & $t_{rel}$ & $r_{rel}$ &$t_{rel}$ & $r_{rel}$ & $t_{rel}$ & $r_{rel}$ & $t_{rel}$ & $r_{rel}$ & \multicolumn{2}{c}{} \\
    \midrule
    \multicolumn{2}{l|}{  Attention-based \cite{liu2021swin}} & \bf{0.44}  & \bf{0.34}  & 1.14  & 0.57  & 1.25  & 0.48  & 1.00  & 0.49  & 0.96  & 0.47  & \multicolumn{2}{c}{183.76 ms} \\
    \multicolumn{2}{l|}{  Convolution-based \cite{huang2020epnet}} & 0.50  & 0.37  & 1.25  & 0.50  & 1.14  & 0.58  & 1.06  & 0.56  & 0.99  & 0.50  & \multicolumn{2}{c}{\bf{87.24 ms}} \\
    \multicolumn{2}{l|}{  Clustering-based} & 0.46  & \bf{0.34}  & \bf{1.09}  & \bf{0.44}  & \bf{0.85}  & \bf{0.36}  & \bf{0.88}  & \bf{0.48}  & \bf{0.82}  & \bf{0.41}  & \multicolumn{2}{c}{98.50 ms} \\
    \bottomrule
    \end{tabular}}
  \label{tab:local stratrgy}%
\end{table}%

\noindent\textbf{Local Fusion Strategy.} We compare the performance of different local fusion strategies. The results in \cref{tab:local stratrgy} show that our clustering-based local fusion strategy outperforms the convolution-based \cite{huang2020epnet} and attention-based \cite{liu2021swin} strategy in accuracy. Also, clustering-based fusion has a gratifying efficiency, where the total inference time is slightly higher than convolution-based fusion and half of attention-based fusion methods.

\section{Conclusion}
In this paper, we propose a novel local-to-global fusion network with the bi-directional structure alignment for visual-LiDAR odometry. A clustering-based local fusion module is designed to provide fine-grained multi-modal feature exchange. Furthermore, an adaptive global fusion is designed to achieve global information interaction. Comprehensive experiments show that our DVLO achieves state-of-the-art performance in terms of both accuracy and efficiency. Our fusion module can also serve as a rather generic fusion strategy, which generalizes well onto the multi-modal scene flow estimation task. We leave the generalization experiments on more multi-modal tasks for future work.

\section*{Acknowledgements}
This work was supported in part by the Natural Science Foundation of China under Grant 62225309, 62073222, 62361166632, and U21A20480.
%
%

\bibliographystyle{splncs04}
\bibliography{main}
\end{document}